\documentclass[sigconf]{acmart}

\AtBeginDocument{%
  \providecommand\BibTeX{{%
    \normalfont B\kern-0.5em{\scshape i\kern-0.25em b}\kern-0.8em\TeX}}}

\copyrightyear{2022} 
\acmYear{2022} 
\setcopyright{acmcopyright}\acmConference[WWW '22 Companion]{Companion Proceedings of the Web Conference 2022}{April 25--29, 2022}{Virtual Event, Lyon, France}
\acmBooktitle{Companion Proceedings of the Web Conference 2022 (WWW '22 Companion), April 25--29, 2022, Virtual Event, Lyon, France}
\acmPrice{15.00}
\acmDOI{10.1145/3487553.3524251}
\acmISBN{978-1-4503-9130-6/22/04}

\usepackage{subcaption}
\usepackage{caption}
\usepackage{multirow}
\usepackage{pgfplots}
\usepackage{pgfplotstable} 
\pgfplotsset{compat=newest}
\usepackage{enumitem}
\usepackage{arydshln}

\begin{document}       

\title{QuatRE: Relation-Aware Quaternions for Knowledge Graph Embeddings}

\author{Dai Quoc Nguyen}
\authornote{This work was done when Dai Quoc Nguyen was a PhD student at Monash University, Australia.}
\affiliation{\country{Oracle Labs, Australia}}
\email{dai.nguyen@oracle.com}

\author{Thanh Vu}
\affiliation{\country{AEHRC, CSIRO, Australia}}
\email{thanh.vu@csiro.au}

\author{Tu Dinh Nguyen}
\affiliation{\country{VinAI Research, Vietnam}}
\email{v.tund21@vinai.io}

\author{Dinh Phung}
\affiliation{\country{Monash University, Australia}}
\email{dinh.phung@monash.edu}

\begin{abstract}

We propose a simple yet effective embedding model to learn quaternion embeddings for entities and relations in knowledge graphs. Our model aims to enhance correlations between head and tail entities given a relation within the Quaternion space with Hamilton product. The model achieves this goal by further associating each relation with two relation-aware rotations, which are used to rotate quaternion embeddings of the head and tail entities, respectively. Experimental results show that our proposed model produces state-of-the-art performances on well-known benchmark datasets for knowledge graph completion.
Our code is available at: \url{https://github.com/daiquocnguyen/QuatRE}.

\end{abstract}



\maketitle

\section{Introduction}

Knowledge graphs (KGs) are constructed to represent relationships between entities in the form of triples \textit{(head, relation, tail)} denoted as \textit{(h, r, t)}.
A typical problem in KGs is the lack of many valid triples \citep{West:2014}; therefore, research approaches have been proposed to predict whether a new triple missed in KGs is likely valid \citep{bordes2011learning,NIPS2013_5071,NIPS2013_5028,NGUYEN2021Thesis,Nguyen2022NoGE}.
These approaches often utilize embedding models to compute a score for each triple, such that valid triples have higher scores than invalid ones.
For example, the score of the valid triple (Melbourne, city\_Of, Australia) is higher than the score of the invalid one (Melbourne, city\_Of, Germany).

Most of the existing models focus on embedding entities and relations within the real-valued vector space \citep{NIPS2013_5071,AAAI148531,AAAI159571,Yang2015,Dettmers2017,Nguyen2018ConvKBfull,Nguyen2020RMeN}.
Moving beyond the real-valued vector space, ComplEx \citep{Trouillon2016} and RotatE \citep{sun2018rotate} consider the complex vector space, MuRP \citep{balazevic2019multi} leverages the hyperbolic space, and QuatE \citep {zhang2019quaternion} learns entity and relation embeddings within the Quaternion space.
However, these existing hyper-complex embedding models just utilize the embedding $\boldsymbol{v}_h$ of the head entity, the embedding $\boldsymbol{v}_r$ of the relation, and the embedding $\boldsymbol{v}_t$ of the tail entity to compute the triple score. 
Therefore, they are not completely effective at capturing the correlations between the head and tail entities. 
For example, given a relation “has positive test'”, the models do not capture fully the correlations between the attributes (e.g., age, gender, and medical record) of the head entity (e.g., “Donald Trump”) and the attributes (e.g., transmission rate and clinical characteristics) of the tail entity (e.g., “COVID-19”). 
Some early translation-based models such as TransR \citep{AAAI159571} and STransE \citep{NguyenNAACL2016} can partially address the issue by associating each relation with translation matrices, but growing model parameters significantly.

Addressing these problems, we propose a simple yet effective embedding model, named {QuatRE}, to learn the quaternion embeddings for entities and relations. 
QuatRE further utilizes two relation-aware rotations for the head and tail embeddings through the Hamilton product, respectively.
QuatRE simplifies the typical use of translation matrices in translation-based models into two quaternion vectors, hence significantly reducing computation.
As a result, QuatRE strengthens the correlations between the head and tail entities.
Experimental results demonstrate that our {QuatRE} obtains state-of-the-art performances on well-known benchmark datasets 
(consisting of WN18, WN18RR, FB15K, and FB15k237) 
for the knowledge graph completion task; thus, it can act as a new strong baseline for future work.

\section{The approach}

\subsection{Quaternion background}
\label{sec:quabackground}


A quaternion $q \in \mathbb{H}$ is a hyper-complex number consisting of a real and three separate imaginary components \citep{hamilton1844ii} defined as:
$q = q_\mathsf{r} + q_\mathsf{i}\boldsymbol{\mathsf{i}} + q_\mathsf{j}\boldsymbol{\mathsf{j}} + q_\mathsf{k}\boldsymbol{\mathsf{k}}$,
where $q_\mathsf{r}, q_\mathsf{i}, q_\mathsf{j}, q_\mathsf{k} \in \mathbb{R}$, and $\boldsymbol{\mathsf{i}}, \boldsymbol{\mathsf{j}}, \boldsymbol{\mathsf{k}}$ are imaginary units that $\boldsymbol{\mathsf{i}}\boldsymbol{\mathsf{j}}\boldsymbol{\mathsf{k}} = \boldsymbol{\mathsf{i}}^2 = \boldsymbol{\mathsf{j}}^2 = \boldsymbol{\mathsf{k}}^2 = -1$, leads to noncommutative multiplication rules as $\boldsymbol{\mathsf{i}}\boldsymbol{\mathsf{j}} = \boldsymbol{\mathsf{k}}, \boldsymbol{\mathsf{j}}\boldsymbol{\mathsf{i}} = -\boldsymbol{\mathsf{k}}, \boldsymbol{\mathsf{j}}\boldsymbol{\mathsf{k}} = \boldsymbol{\mathsf{i}}, \boldsymbol{\mathsf{k}}\boldsymbol{\mathsf{j}} = -\boldsymbol{\mathsf{i}}, \boldsymbol{\mathsf{k}}\boldsymbol{\mathsf{i}} = \boldsymbol{\mathsf{j}}$, and $\boldsymbol{\mathsf{i}}\boldsymbol{\mathsf{k}} = -\boldsymbol{\mathsf{j}}$.
Correspondingly, a $n$-dimensional quaternion vector $\boldsymbol{q} \in \mathbb{H}^n$ is defined as:
$
\boldsymbol{q} = \boldsymbol{q}_\mathsf{r} + \boldsymbol{q}_\mathsf{i}\boldsymbol{\mathsf{i}} + \boldsymbol{q}_\mathsf{j}\boldsymbol{\mathsf{j}} + \boldsymbol{q}_\mathsf{k}\boldsymbol{\mathsf{k}} 
$,
where $\boldsymbol{q}_\mathsf{r}, \boldsymbol{q}_\mathsf{i}, \boldsymbol{q}_\mathsf{j}, \boldsymbol{q}_\mathsf{k} \in \mathbb{R}^n$.

\paragraph{Norm.} 
The normalized quaternion vector $\boldsymbol{q}^\triangleleft$ of $\boldsymbol{q} \in \mathbb{H}^n$ is computed as:
$\boldsymbol{q}^\triangleleft = \frac{\boldsymbol{q}_\mathsf{r} + \boldsymbol{q}_\mathsf{i}\boldsymbol{\mathsf{i}} + \boldsymbol{q}_\mathsf{j}\boldsymbol{\mathsf{j}} + \boldsymbol{q}_\mathsf{k}\boldsymbol{\mathsf{k}}}{\sqrt{\boldsymbol{q}_\mathsf{r}^2 + \boldsymbol{q}_\mathsf{i}^2 + \boldsymbol{q}_\mathsf{j}^2 + \boldsymbol{q}_\mathsf{k}^2}}$

\paragraph{Hamilton product.} 
The Hamilton product of two vectors $\boldsymbol{q}$ and $\boldsymbol{p} \in \mathbb{H}^n$ is computed as:
\begin{eqnarray}
\boldsymbol{q} \otimes \boldsymbol{p} &=& (\boldsymbol{q}_\mathsf{r} \circ \boldsymbol{p}_\mathsf{r} - \boldsymbol{q}_\mathsf{i} \circ \boldsymbol{p}_\mathsf{i} - \boldsymbol{q}_\mathsf{j} \circ \boldsymbol{p}_\mathsf{j} - \boldsymbol{q}_\mathsf{k} \circ \boldsymbol{p}_\mathsf{k})  \nonumber \\
&+& (\boldsymbol{q}_\mathsf{i} \circ \boldsymbol{p}_\mathsf{r} + \boldsymbol{q}_\mathsf{r} \circ \boldsymbol{p}_\mathsf{i} - \boldsymbol{q}_\mathsf{k} \circ \boldsymbol{p}_\mathsf{j} + \boldsymbol{q}_\mathsf{j} \circ \boldsymbol{p}_\mathsf{k})\boldsymbol{\mathsf{i}} \nonumber \\
&+& (\boldsymbol{q}_\mathsf{j} \circ \boldsymbol{p}_\mathsf{r} + \boldsymbol{q}_\mathsf{k} \circ \boldsymbol{p}_\mathsf{i} + \boldsymbol{q}_\mathsf{r} \circ \boldsymbol{p}_\mathsf{j} - \boldsymbol{q}_\mathsf{i} \circ \boldsymbol{p}_\mathsf{k})\boldsymbol{\mathsf{j}} \nonumber \\
&+& (\boldsymbol{q}_\mathsf{k} \circ \boldsymbol{p}_\mathsf{r} - \boldsymbol{q}_\mathsf{j} \circ \boldsymbol{p}_\mathsf{i} + \boldsymbol{q}_\mathsf{i} \circ \boldsymbol{p}_\mathsf{j} + \boldsymbol{q}_\mathsf{r} \circ \boldsymbol{p}_\mathsf{k})\boldsymbol{\mathsf{k}} \nonumber
\label{equa:halproduct}
\end{eqnarray}
where $\circ$ denotes the element-wise product.
We note that the Hamilton product is not commutative, i.e., $q \otimes p \neq p \otimes q$.

\paragraph{Quaternion-inner product.}  The quaternion-inner product $\bullet$ of two quaternion vectors $\boldsymbol{q}$ and $\boldsymbol{p} \in \mathbb{H}^n$ returns a scalar, which is computed as:
$\boldsymbol{q} \bullet \boldsymbol{p} = \boldsymbol{q}_\mathsf{r}^\textsf{T}\boldsymbol{p}_\mathsf{r} + \boldsymbol{q}_\mathsf{i}^\textsf{T}\boldsymbol{p}_\mathsf{i} + \boldsymbol{q}_\mathsf{j}^\textsf{T}\boldsymbol{p}_\mathsf{j} + \boldsymbol{q}_\mathsf{k}^\textsf{T}\boldsymbol{p}_\mathsf{k} $

\subsection{The proposed QuatRE}
\label{sec:ourmodel}

A knowledge graph (KG) $\mathcal{G}$ is a collection of valid factual triples in the form of \textit{(head, relation, tail)} denoted as $(h, r, t)$ such that $h, t \in \mathcal{E}$ and $r \in \mathcal{R}$ where $\mathcal{E}$ is a set of entities and $\mathcal{R}$ is a set of relations.
KG embedding models aim to embed entities and relations to a low-dimensional vector space to define a score function $f$. This function is to give a score for each triple $(h, r, t)$, such that the valid triples obtain higher scores than the invalid triples.

The existing hyper-complex embedding models, such as ComplEx, RotatE, and QuatE, only utilize $\boldsymbol{v}_h$, $\boldsymbol{v}_r$, $\boldsymbol{v}_t$ to obtain the triple score; hence they are not completely effective at modeling the correlations between the head and tail entities.
For example, given a relation “has positive test'”, these models do not capture fully the correlations between the attributes (e.g., age, gender, and medical record) of the head entity (e.g., “Donald Trump”) and the attributes (e.g., transmission rate and clinical characteristics) of the tail entity (e.g., “COVID-19”).
Therefore, we propose QuatRE, a simple yet effective KG embedding model, to overcome this limitation by integrating relation-aware rotations to increase the correlations between the head and tail entities.

Given a triple $(h, r, t)$, QuatRE also represents the embeddings of entities and relations within the Quaternion space.
The quaternion embeddings  $\boldsymbol{v}_h$, $\boldsymbol{v}_r$, and $\boldsymbol{v}_t \in \mathbb{H}^n$ of $h$, $r$, and $t$ are represented as:
\begin{eqnarray}
\boldsymbol{v}_h &=& \boldsymbol{v}_{h,\mathsf{r}} + \boldsymbol{v}_{h,\mathsf{i}}\boldsymbol{\mathsf{i}} + \boldsymbol{v}_{h,\mathsf{j}}\boldsymbol{\mathsf{j}} + \boldsymbol{v}_{h,\mathsf{k}}\boldsymbol{\mathsf{k}} \\
\boldsymbol{v}_r &=& \boldsymbol{v}_{r,\mathsf{r}} + \boldsymbol{v}_{r,\mathsf{i}}\boldsymbol{\mathsf{i}} + \boldsymbol{v}_{r,\mathsf{j}}\boldsymbol{\mathsf{j}} + \boldsymbol{v}_{r,\mathsf{k}}\boldsymbol{\mathsf{k}} \\
\boldsymbol{v}_t &=& \boldsymbol{v}_{t,\mathsf{r}} + \boldsymbol{v}_{t,\mathsf{i}}\boldsymbol{\mathsf{i}} + \boldsymbol{v}_{t,\mathsf{j}}\boldsymbol{\mathsf{j}} + \boldsymbol{v}_{t,\mathsf{k}}\boldsymbol{\mathsf{k}}
\end{eqnarray}
where $\boldsymbol{v}_{h,\mathsf{r}}$, $\boldsymbol{v}_{h,\mathsf{i}}$, $\boldsymbol{v}_{h,\mathsf{j}}$, $\boldsymbol{v}_{h,\mathsf{k}}$, $\boldsymbol{v}_{r,\mathsf{r}}$, $\boldsymbol{v}_{r,\mathsf{i}}$, $\boldsymbol{v}_{r,\mathsf{j}}$, $\boldsymbol{v}_{r,\mathsf{k}}$, $\boldsymbol{v}_{t,\mathsf{r}}$, $\boldsymbol{v}_{t,\mathsf{i}}$, $\boldsymbol{v}_{t,\mathsf{j}}$, and $\boldsymbol{v}_{t,\mathsf{k}} \in \mathbb{R}^n$.
QuatRE further associates each relation $r$ with two quaternion vectors $\boldsymbol{\mathsf{v}}_{r,1}$ and $\boldsymbol{\mathsf{v}}_{r,2} \in \mathbb{H}^n$ as:
\begin{eqnarray}
\boldsymbol{\mathsf{v}}_{r,1} &=& \boldsymbol{\mathsf{v}}_{r,1,\mathsf{r}} + \boldsymbol{\mathsf{v}}_{r,1,\mathsf{i}}\boldsymbol{\mathsf{i}} + \boldsymbol{\mathsf{v}}_{r,1,\mathsf{j}}\boldsymbol{\mathsf{j}} + \boldsymbol{\mathsf{v}}_{r,1,\mathsf{k}}\boldsymbol{\mathsf{k}} \\
\boldsymbol{\mathsf{v}}_{r,2} &=& \boldsymbol{\mathsf{v}}_{r,2,\mathsf{r}} + \boldsymbol{\mathsf{v}}_{r,2,\mathsf{i}}\boldsymbol{\mathsf{i}} + \boldsymbol{\mathsf{v}}_{r,2,\mathsf{j}}\boldsymbol{\mathsf{j}} + \boldsymbol{\mathsf{v}}_{r,2,\mathsf{k}}\boldsymbol{\mathsf{k}}
\end{eqnarray}
where $\boldsymbol{\mathsf{v}}_{r,1,\mathsf{r}}$, $\boldsymbol{\mathsf{v}}_{r,1,\mathsf{i}}$, $\boldsymbol{\mathsf{v}}_{r,1,\mathsf{j}}$, $\boldsymbol{\mathsf{v}}_{r,1,\mathsf{k}}$, $\boldsymbol{\mathsf{v}}_{r,2,\mathsf{r}}$, $\boldsymbol{\mathsf{v}}_{r,2,\mathsf{i}}$,  $\boldsymbol{\mathsf{v}}_{r,2,\mathsf{j}}$, and $\boldsymbol{\mathsf{v}}_{r,2,\mathsf{k}} \in \mathbb{R}^n$.
QuatRE then uses the Hamilton product to rotate $\boldsymbol{v}_h$ and $\boldsymbol{v}_t$ by the normalized vectors $\boldsymbol{\mathsf{v}}_{r,1}^\triangleleft$ and $\boldsymbol{\mathsf{v}}_{r,2}^\triangleleft$ respectively as:
\begin{eqnarray}
\boldsymbol{v}_{h,r,1} &=& \boldsymbol{v}_h \otimes \boldsymbol{\mathsf{v}}_{r,1}^\triangleleft \label{equa:rotate_head}\\
\boldsymbol{v}_{t,r,2} &=& \boldsymbol{v}_t \otimes \boldsymbol{\mathsf{v}}_{r,2}^\triangleleft \label{equa:rotate_tail}
\end{eqnarray}
After that, QuatRE also utilizes a Hamilton product-based rotation for $\boldsymbol{v}_{h,r,1}$ by the normalized quaternion embedding $\boldsymbol{v}_r^\triangleleft$, then followed by a quaternion-inner product with $\boldsymbol{v}_{t,r,2}$ to produce the triple score.
The quaternion components of input vectors are shared during computing the Hamilton product, as shown in Equation \ref{equa:halproduct}.
Therefore, QuatRE uses two rotations in Equations \ref{equa:rotate_head} and \ref{equa:rotate_tail} for $\boldsymbol{v}_h$ and $\boldsymbol{v}_t$ to increase the correlations between the head $h$ and tail $t$ entities given the relation $r$.


Formally, we define the QuatRE score function $f$ for the triple $(h, r, t)$ as:
\begin{eqnarray}
f(h, r, t) &=& \left(\boldsymbol{v}_{h,r,1} \otimes  \boldsymbol{v}_r^\triangleleft\right) \bullet \boldsymbol{v}_{t,r,2} 
\nonumber \\
&=&
\left(\left(\boldsymbol{v}_h \otimes \boldsymbol{\mathsf{v}}_{r,1}^\triangleleft\right) \otimes  \boldsymbol{v}_r^\triangleleft\right) \bullet \left(\boldsymbol{v}_t \otimes \boldsymbol{\mathsf{v}}_{r,2}^\triangleleft\right)
\end{eqnarray}


\paragraph{Proposition.} If we fix the real components of both $\boldsymbol{\mathsf{v}}_{r,1}$ and $\boldsymbol{\mathsf{v}}_{r,2}$ to \textbf{1}, and fix the imaginary components of both $\boldsymbol{\mathsf{v}}_{r,1}$ and $\boldsymbol{\mathsf{v}}_{r,2}$ to \textbf{0}, our QuatRE is simplified to QuatE. 
Hence QuatRE is viewed as an extension of QuatE. 
Furthermore, given the same embedding dimension $n$, QuatE has $(|\mathcal{E}|\times 4\times n + |\mathcal{R}|\times 4\times n)$ parameters, while QuatRE has $(|\mathcal{E}|\times 4\times n + 3\times |\mathcal{R}|\times 4\times n)$ parameters. Given that  $|\mathcal{R}|$ is significantly smaller than $|\mathcal{E}|$; hence QuatE and our QuatRE have comparable numbers of parameters.
Besides, an advantage of QuatRE is to change the common use of translation matrices in translation-based models such as TransR \citep{AAAI159571} and STransE \citep{NguyenNAACL2016}, hence reducing computation significantly.

\paragraph{Learning process.} We employ the Adagrad optimizer \citep{duchi2011adaptive} to train our proposed QuatRE by minimizing the following loss function \citep{Trouillon2016} with the regularization on model parameters $\boldsymbol{\theta}$ as:

\begin{eqnarray}
\mathcal{L} =  \sum_{\substack{(h, r, t) \in \{\mathcal{G} \cup \mathcal{G}'\}}} &\log&\left(1 + \exp\left(- l_{(h, r, t)} \cdot f(h, r, t)\right)\right)
+
\lambda\|\boldsymbol{\theta}\|_2^2 \nonumber \\
\label{equal:objfunc}
\end{eqnarray}
\begin{equation*}
\text{in which, } l_{(h, r, t)} = \left\{ 
  \begin{array}{l}
  1 \ \ \ \text{for } (h, r, t)\in\mathcal{G}\\
 -1 \ \ \ \text{for } (h, r, t)\in\mathcal{G}'
  \end{array} \right.
\end{equation*}
where we use $l_2$-norm with the regularization rate $\lambda$; and $\mathcal{G}$ and $\mathcal{G}'$ are collections of valid and invalid triples, respectively.  
$\mathcal{G}'$ is generated by corrupting valid triples in $\mathcal{G}$. 

\section{Experimental setup}
\label{ssec:kbc1}

The knowledge graph completion task \citep{NIPS2013_5071} is to predict a missing entity given a relation with another entity, for example, inferring a head entity $h$ given $(r, t)$ or inferring a tail entity $t$ given $(h, r)$. 
The results are calculated by ranking the scores produced by the score function $f$ on triples in the test set.

\paragraph{Datasets} 
We evaluate our proposed QuatRE for the knowledge graph completion task \citep{NIPS2013_5071} on four well-known benchmark datasets: 
WN18, FB15k \citep{NIPS2013_5071}, 
WN18RR \citep{Dettmers2017} and FB15k-237 \citep{toutanova-chen:2015:CVSC}.
As mentioned in \citep{toutanova-chen:2015:CVSC}, WN18 and FB15k contains many reversible relations, which makes the prediction task become  trivial and unrealistic.
Therefore, their subsets WN18RR and FB15k-237 are derived to eliminate the reversible relation problem to create more realistic and challenging prediction tasks. 


\paragraph{Evaluation protocol}
Following \citet{NIPS2013_5071}, for each valid test triple $(h, r, t)$, we replace either $h$ or $t$ by each of other entities to create a set of corrupted triples.
We use the ``{Filtered}'' setting protocol \citep{NIPS2013_5071}, i.e., not including any corrupted triples that appear in the KG.
We rank the valid test triple and corrupted triples in descending order of their scores.
We employ evaluation metrics: mean rank (MR), mean reciprocal rank (MRR), and Hits@$k$.
The final scores on the test set are reported for the model which obtains the highest Hits@10 on the validation set.
We follow \citep{zhang2019quaternion} to report two versions of our QuatRE for a fair comparison with QuatE.


\paragraph{Training protocol}


We set 100 batches for all datasets. We then vary the learning rate $\alpha$ in \{0.02, 0.05, 0.1\}, the number $s$ of negative triples sampled per training triple in \{1, 5, 10\}, the embedding dimension $n$ in \{128, 256, 384\}, and the regularization rate $\lambda$ in $\{0.05, 0.1, 0.2, 0.5\}$.
We train our QuatRE up to 8,000 epochs on 
WN18 and 
WN18RR and 2,000 epochs on 
FB15k and 
FB15k-237. 
We monitor the Hits@10 score after each 400 epochs on
WN18 and 
WN18RR and each 200 epochs on 
FB15k and 
FB15k-237.
We select the hyper-parameters using grid search and early stopping on the validation set with Hits@10.

\section{Experimental results}
\label{ssec:kbc}


\begin{table}[!htb]
\centering
\caption{Experimental results on WN18 and FB15k. Hits@$k$ (H@$k$) is reported in \%. 
The best scores are in {bold}, while the second best scores are in {underline}.
}
\resizebox{8.5cm}{!}{
\setlength{\tabcolsep}{0.2em}
\begin{tabular}{l|ccccc|ccccc}
\hline 
\multirow{2}{*}{\bf Method}& \multicolumn{5}{c|}{\bf WN18} & \multicolumn{5}{c}{\bf FB15k}\\
\cline{2-11}
 & MR & MRR & H@10 & H@3 & H@1 & MR & MRR & H@10 & H@3 & H@1\\
\hline
\hline
TransE \citep{NIPS2013_5071} & -- & 0.495 & 94.3 & 88.8 & 11.3 & -- & 0.463 & 74.9 & 57.8 & 29.7 \\
STransE \citep{NguyenNAACL2016} & 206 & 0.657 & 93.4 & -- & -- & 69 & 0.543 & 79.7 & -- & --\\
DistMult \citep{Yang2015}& 655 & 0.797 & 94.6 & -- & -- & 42 & \underline{0.798} & 89.3 & -- & --\\
ConvE \citep{Dettmers2017} & 374 & 0.943 & 95.6 & 94.6 & 93.5 & 51 & 0.657 & 83.1 & 72.3 & 55.8\\
\hline
ComplEx \citep{Trouillon2016} & -- & 0.941 & 94.7 & 94.5 & 93.6 & -- & 0.692 & 84.0 & 75.9 & 59.9\\
TorusE \citep{ebisu2018toruse} & -- & \underline{0.947} & 95.4 & 95.0 & \underline{94.3} & -- & 0.733 & 83.2 & 77.1 & 67.4\\
RotatE \citep{sun2018rotate} & 184 & \underline{0.947} & \underline{96.1} & \underline{95.3} & 93.8 & 32 & 0.699 & 87.2 & 78.8 & 58.5\\
QuatE$_{1}$ \citep{zhang2019quaternion} & 388 & {0.949} & 96.0 & \textbf{95.4} & 94.1 & 41 & 0.770 & 87.8 & 82.1 & 70.0\\
QuatE$_{2}$ \citep{zhang2019quaternion} & \underline{162} & \textbf{0.950} & 95.9 & \textbf{95.4} & \textbf{94.5} & \textbf{17} & 0.782 & \textbf{90.0} & \underline{83.5} & \underline{71.1}\\
\hline
\textbf{QuatRE}$_{1}$ & 249 & 0.936 & \underline{96.1} & 95.1 & 91.9 & 44 & {0.786} & 88.1 & 83.0 & {72.5}\\ 
\textbf{QuatRE}$_{2}$ & \textbf{116} & 0.939 & \textbf{96.3} & \underline{95.3} & 92.3 & \underline{23} & \textbf{0.808} & \underline{89.6} & \textbf{85.1} & \textbf{75.1}\\
\hline
\end{tabular}
}
\label{tab:resultswn18fb15k}
\end{table}

\paragraph{Main results}

We report the experimental results on the datasets in Tables \ref{tab:resultswn18fb15k} and 
\ref{tab:resultswn18rrfb15k237}.
Our proposed QuatRE produces competitive results compared to the up-to-date models across all metrics.
QuatRE achieves the best scores for MR and Hits@10 on WN18, and MRR, Hits@3, and Hits@1 on FB15k, and obtains competitive scores for other metrics on these two datasets.
On more challenging datasets WN18RR and FB15k-237, our 
QuatRE outperforms up-to-date baselines for all metrics except the Hits@1 on WN18RR and the second-best MR on FB15k-237.
Especially when comparing with QuatE, on WN18RR, QuatRE gains significant improvements of $2314-1986=328$ in MR (which is about 14\% relative improvement), and 1.0\% and 1.1\% absolute improvements in Hits@10 and Hits@3 respectively. 
Besides, on FB15k-237, QuatRE achieves improvements of $0.367-0.348=0.019$ in MRR (which is 5.5\% relative improvement) and obtains absolute gains of 1.3\%, 2.2\%, and 2.1\% in Hits@10, Hits@3, and Hits@1 respectively.

\begin{table}[!htb]
\centering
\caption{Experimental results on WN18RR and FB15k-237. Hits@$k$ (H@$k$) is reported in \%. 
The best scores are in {bold}, while the second best scores are in {underline}.
The results of TransE are taken from \citep{Nguyen2018}.
The results of DistMult and ComplEx are taken from \citep{Dettmers2017}.
}
\resizebox{8.5cm}{!}{
\setlength{\tabcolsep}{0.2em}
\begin{tabular}{l|ccccc|ccccc}
\hline
\multirow{2}{*}{\bf Method}& \multicolumn{5}{c|}{\bf WN18RR} & \multicolumn{5}{c}{\bf FB15k-237}\\
\cline{2-11}
 & MR & MRR & H@10 & H@3 & H@1 & MR & MRR & H@10 & H@3 & H@1\\
\hline
TransE \citep{NIPS2013_5071} & 3384 & 0.226 & 50.1 & -- & -- & 357 & 0.294 & 46.5 & -- & --\\
DistMult \citep{Yang2015} & 5110 & 0.430 & 49.0 & 44.0 & 39.0 & 254 & 0.241 & 41.9 & 26.3 & 15.5\\
ConvE \citep{Dettmers2017} & 5277 & 0.460 & 48.0 & 43.0 & 39.0 & 246 & {0.316} & 49.1 & {35.0} & {23.9}\\
ConvKB \citep{Nguyen2018} & {2741} & 0.220 & 50.8 & -- & -- & 196 & 0.302 & 48.3 & -- & -- \\ 
AutoSF \citep{ZhangAutoSF} & -- & \underline{0.490} & 56.7 & -- & \textbf{45.1} & -- & \underline{0.360} & \underline{55.2} & -- & \underline{26.7}\\
\hline
ComplEx \citep{Trouillon2016} & 5261 & 0.440 & 51.0 & 46.0 & 41.0 & 339 & 0.247 & 42.8 & 27.5 & 15.8\\
RotatE \citep{sun2018rotate} & 3277 & 0.470 & 56.5 & 48.8 & 42.2 & 185 & 0.297 & 48.0 & 32.8 & 20.5\\
MuRP \citep{balazevic2019multi} & -- & 0.481 & 56.6 & 49.5 & \underline{44.0} & -- & 0.335 & 51.8 & 36.7 & 24.3 \\
QuatE$_{1}$ \citep{zhang2019quaternion} & 3472 & 0.481 & 56.4 & 50.0 & 43.6 & 176 & 0.311 & 49.5 & 34.2 & 22.1\\
QuatE$_{2}$ \citep{zhang2019quaternion} & \underline{2314} & {0.488} & \underline{58.2} & \underline{50.8} & {43.8} & \textbf{87} & {0.348} & {55.0} & \underline{38.2} & {24.8}\\
\hline
\textbf{QuatRE}$_{1}$ & 3038 & 0.479 & 57.1 & 50.3 & 42.9 & 168 & 0.332 & 52.2 & 36.7 & 23.8\\
\textbf{QuatRE}$_{2}$ & \textbf{1986} & \textbf{0.493} & \textbf{59.2} & \textbf{51.9} & {43.9} & \underline{88} & \textbf{0.367} & \textbf{56.3} & \textbf{40.4} & \textbf{26.9} \\
\hline
\end{tabular}
}
\label{tab:resultswn18rrfb15k237}
\end{table}

\paragraph{Correlation analysis} 

\begin{figure}[!ht]
\centering
    \includegraphics[width=0.225\textwidth]{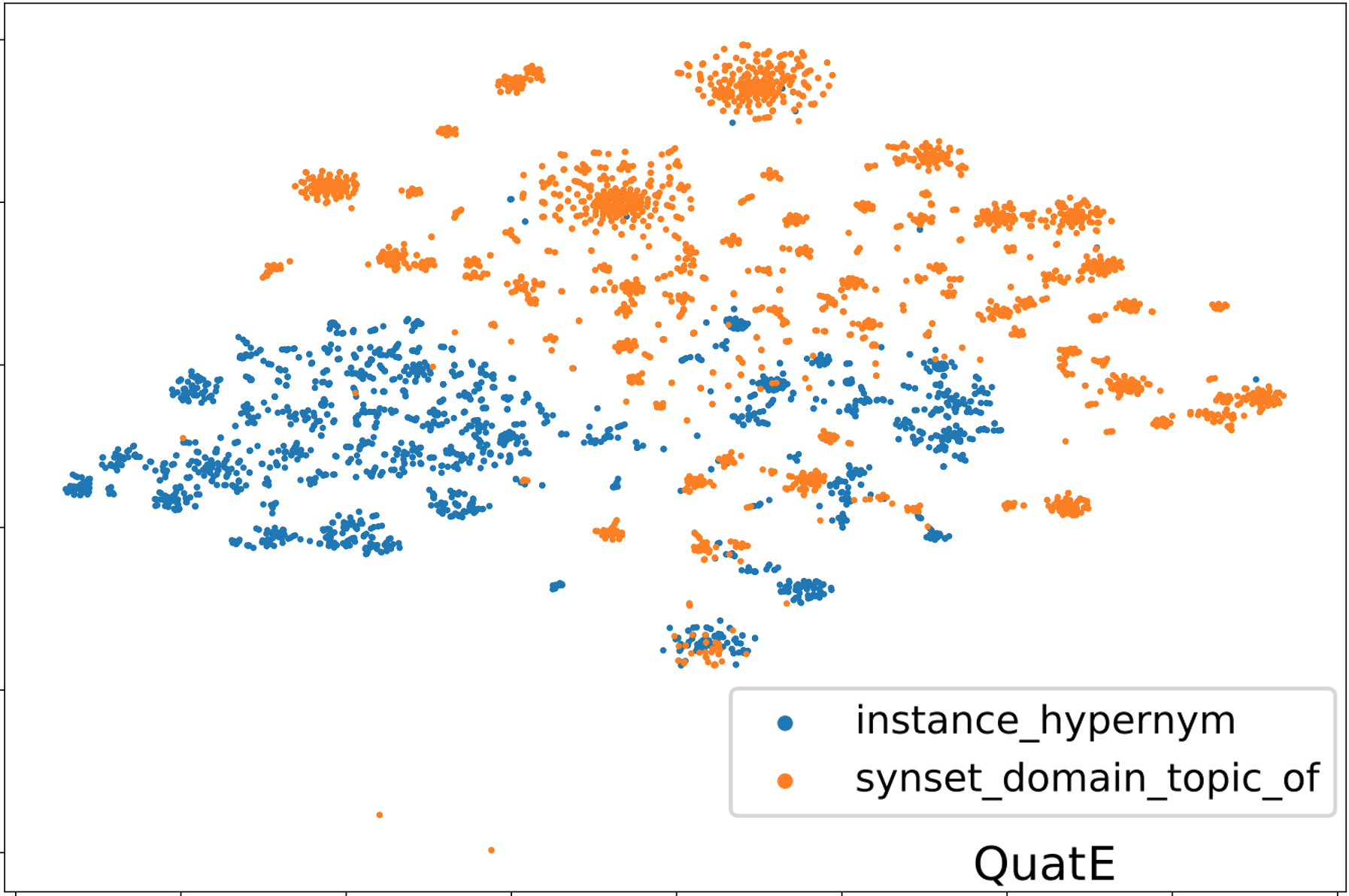}
    \includegraphics[width=0.225\textwidth]{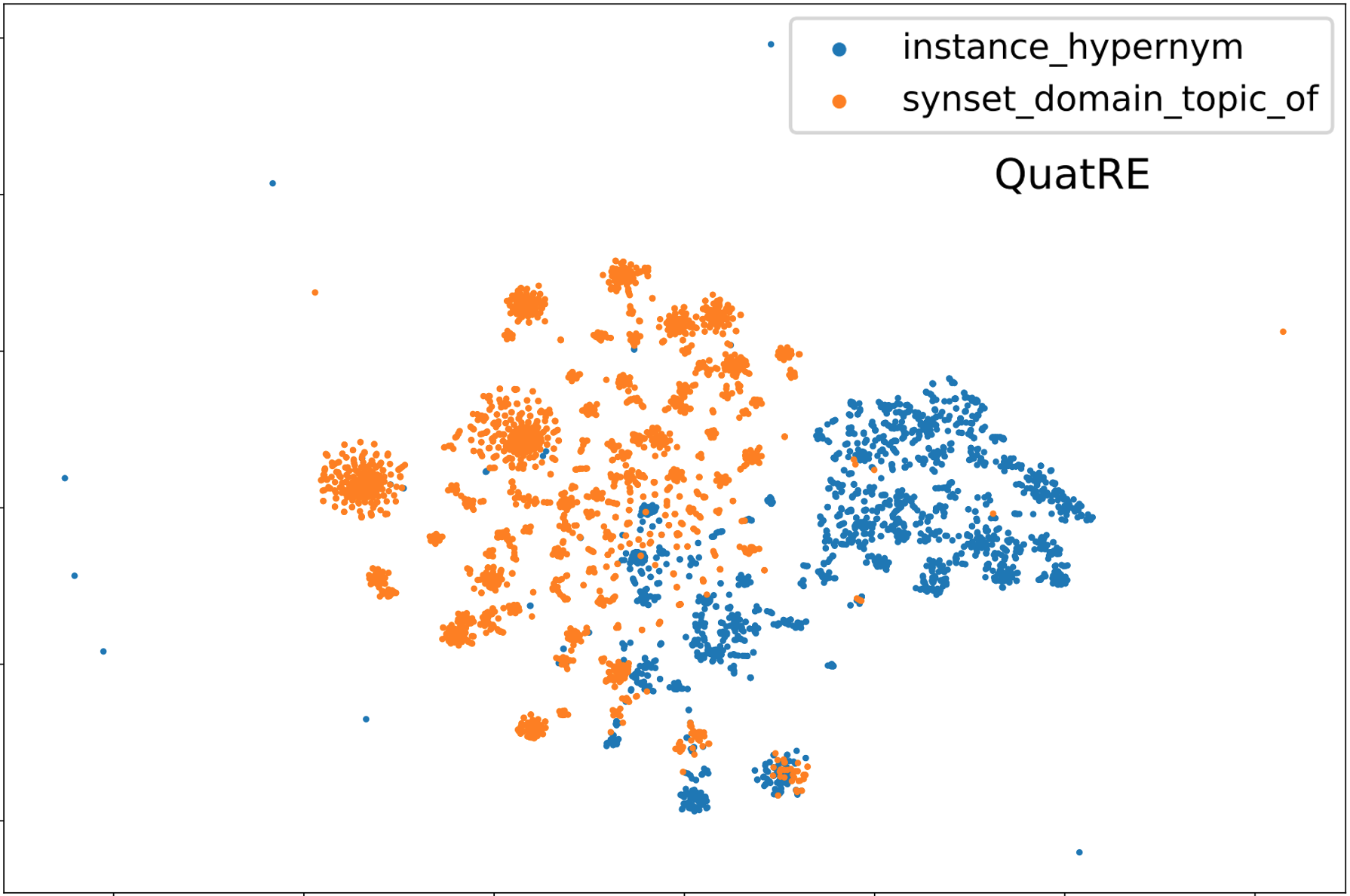}
\caption{A visualization of the learned entity embeddings on WN18RR.}
\label{fig:visualization_of_entities}
\end{figure}

We use t-SNE \citep{maaten2008visualizing} to visualize the learned quaternion embeddings of the entities on WN18RR for QuatE and QuatRE.
We select all entities associated with two relations consisting of ``synset\_domain\_topic\_of'' and ``instance\_hypernym''.
We then vectorize each quaternion embedding using a vector concatenation across the four components; hence, we obtain a real-valued vector representation for applying t-SNE.
Figure \ref{fig:visualization_of_entities} qualitatively demonstrates that QuatRE strengthens the correlations between the entities.

\begin{figure}[!ht]
\centering
\begin{minipage}{0.49\textwidth}
\centering
\resizebox{8.5cm}{!}{
\begin{tikzpicture}
\begin{axis}[
	title = Predicting $head$, 
    ybar,
    enlarge x limits=0.25,
    legend style={at={(0.275,1)},
                anchor=north,legend columns=2},
    ylabel={MRR},
    symbolic x coords={1-1, 1-M, M-1, M-M},
    xtick=data,
    ymin=0.1,ymax=.85,
    nodes near coords=\rotatebox{90}{\scriptsize\pgfmathprintnumber\pgfplotspointmeta},
    ]
\addplot coordinates {(1-1,0.404305) (1-M,0.461112) (M-1,0.277062) (M-M,0.25944)};
\addplot coordinates {(1-1,0.444379) (1-M,0.461081) (M-1,0.289128) (M-M,0.264982)};
\legend{QuatE,QuatRE}
\end{axis}
\end{tikzpicture}
\hspace{0.25cm}
\begin{tikzpicture}
\begin{axis}[
	title = Predicting $tail$, 
    ybar,
    enlarge x limits=0.25,
    legend style={at={(0.275,1)},
                anchor=north,legend columns=2},
    ylabel={MRR},
    symbolic x coords={1-1, 1-M, M-1, M-M},
    xtick=data,
    ymin=0.1,ymax=.85,
    nodes near coords=\rotatebox{90}{\scriptsize\pgfmathprintnumber\pgfplotspointmeta},
    ]
\addplot coordinates {(1-1,0.39412) (1-M,0.157659) (M-1,0.769162) (M-M,0.378297)};
\addplot coordinates {(1-1,0.441859) (1-M,0.161259) (M-1,0.777297) (M-M,0.382173)};
\legend{QuatE,QuatRE}
\end{axis}
\end{tikzpicture}
}
\end{minipage}
\begin{minipage}{0.49\textwidth}
\centering
\resizebox{8.5cm}{!}{
\begin{tikzpicture}
\begin{axis}[
	title = Predicting $head$, 
    ybar,
    enlarge x limits=0.25,
    legend style={at={(0.275,1)},
                anchor=north,legend columns=2},
    ylabel={Hits@10},
    symbolic x coords={1-1, 1-M, M-1, M-M},
    xtick=data,
    ymin=20,ymax=97,
    nodes near coords=\rotatebox{90}{\scriptsize\pgfmathprintnumber\pgfplotspointmeta},
    ]
\addplot coordinates {(1-1,54.2) (1-M,66.4) (M-1,38.6) (M-M,46.9)};
\addplot coordinates {(1-1,58.9) (1-M,66.4) (M-1,39.3) (M-M,48.1)};
\legend{QuatE,QuatRE}
\end{axis}
\end{tikzpicture}
\hspace{0.25cm}
\begin{tikzpicture}
\begin{axis}[
	title = Predicting $tail$, 
    ybar,
    enlarge x limits=0.25,
    legend style={at={(0.275,1)},
                anchor=north,legend columns=2},
    ylabel={Hits@10},
    symbolic x coords={1-1, 1-M, M-1, M-M},
    xtick=data,
    ymin=20,ymax=97,
    nodes near coords=\rotatebox{90}{\scriptsize\pgfmathprintnumber\pgfplotspointmeta},
    ]
\addplot coordinates {(1-1,53.1) (1-M,25.5) (M-1,88.3) (M-M,60.9)};
\addplot coordinates {(1-1,59.9) (1-M,26.8) (M-1,88.9) (M-M,61.7)};
\legend{QuatE,QuatRE}
\end{axis}
\end{tikzpicture}
}
\end{minipage}
\caption{MRR and Hits@10 on FB15k-237 for QuatE and our QuatRE with respect to each relation category.}
\label{fig:mrrrelationtype}
\end{figure}
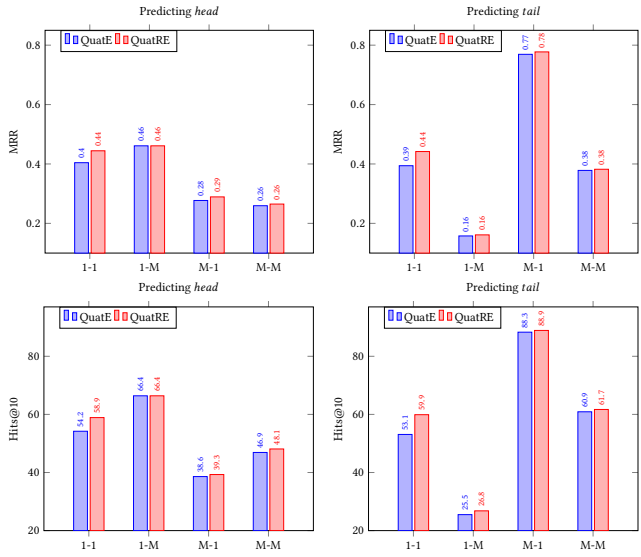

\paragraph{Relation analysis}

Following \citet{NIPS2013_5071}, for each relation $r$, we calculate the averaged number $\eta_h$ of head entities per tail entity and the averaged number $\eta_t$ of tail entities per head entity. 
If $\eta_h < $1.5 and $\eta_t < $1.5, $r$ is categorized  one-to-one  (1-1).
If $\eta_h < $1.5 and $\eta_t \geq $1.5, $r$ is categorized  one-to-many (1-M).
If $\eta_h \geq $1.5 and $\eta_t < $1.5, $r$ is categorized  many-to-one (M-1).
If $\eta_h \geq $1.5 and $\eta_t \geq $1.5, $r$ is categorized  many-to-many (M-M).
Figure \ref{fig:mrrrelationtype} shows the MRR and H@10 scores for predicting the head entities and then the tail entities with respect to each relation category on FB15k-237, wherein our QuatRE outperforms QuatE on these relation categories.
We also report the MRR scores for each relation on WN18RR in Table \ref{tab:mrrwn18rreachrelation}, which shows the effectiveness of QuatRE in modeling different types of relations.

\begin{table}
\centering
\caption{MRR score on the WN18RR test set for each relation.}
\resizebox{5.5cm}{!}{
\begin{tabular}{l|cc}
\hline
{\bf Relation} &  \textbf{QuatE} & \textbf{QuatRE}\\
\hline
hypernym & 0.173 & \textbf{0.190} \\
derivationally\_related\_form & \textbf{0.953} & 0.943 \\
instance\_hypernym & 0.364 & \textbf{0.380} \\
also\_see & 0.629 & \textbf{0.633} \\
member\_meronym & 0.232 & \textbf{0.237} \\
synset\_domain\_topic\_of & 0.468 & \textbf{0.495} \\
has\_part & \textbf{0.233} & 0.226 \\
member\_of\_domain\_usage & 0.441 & \textbf{0.470} \\
member\_of\_domain\_region & 0.193 & \textbf{0.364} \\
verb\_group & \textbf{0.924} & 0.867 \\
similar\_to & \textbf{1.000} & \textbf{1.000} \\
\hline
\end{tabular}
}
\label{tab:mrrwn18rreachrelation}
\end{table}

\paragraph{Ablation analysis}

\begin{table}[!ht]
\centering
\caption{Ablation results. (i) With only using $\boldsymbol{\mathsf{v}}_{r,1}$. (ii) With only using $\boldsymbol{\mathsf{v}}_{r,2}$.
}
\resizebox{8.5cm}{!}{
\setlength{\tabcolsep}{0.6em}
\begin{tabular}{l|cc|cc}
\hline
\multirow{2}{*}{\bf Model}& \multicolumn{2}{c|}{\bf WN18RR} & \multicolumn{2}{c}{\bf FB15k-237} \\
\cline{2-5}
 & MRR & H@10 & MRR & H@10\\
\hline
QuatRE: $\left(\left(\boldsymbol{v}_h \otimes \boldsymbol{\mathsf{v}}_{r,1}^\triangleleft\right) \otimes  \boldsymbol{v}_r^\triangleleft\right) \bullet \left(\boldsymbol{v}_t \otimes \boldsymbol{\mathsf{v}}_{r,2}^\triangleleft\right)$ & \bf 0.493 & \bf  59.2 & \bf 0.367 & \bf 56.3\\
\hdashline
\ \ \ \ \ \ \ \ \ (i) $\left(\left(\boldsymbol{v}_h \otimes \boldsymbol{\mathsf{v}}_{r,1}^\triangleleft\right) \otimes  \boldsymbol{v}_r^\triangleleft\right) \bullet \boldsymbol{v}_t$ & \underline{0.491} & \underline{58.9} & \underline{0.364} & 56.0\\
\hdashline
\ \ \ \ \ \ \ \ \ (ii) $\left(\boldsymbol{v}_h \otimes \boldsymbol{v}_r^\triangleleft\right) \bullet \left(\boldsymbol{v}_t \otimes \boldsymbol{\mathsf{v}}_{r,2}^\triangleleft\right)$ & \underline{0.491} & 58.8 & \underline{0.364} & \underline{56.1}\\
\hline
\hline
{QuatE:} $\left(\boldsymbol{v}_h \otimes  \boldsymbol{v}_r^\triangleleft\right) \bullet \boldsymbol{v}_t$ & 0.488 & 58.2 & 0.348 & 55.0\\
\hline
\end{tabular}
}
\label{tab:ablationstudies}
\end{table}

We report our ablation results for two variants of our QuatRE in Table \ref{tab:ablationstudies}, wherein we only use either $\boldsymbol{\mathsf{v}}_{r,1}$ to rotate $\boldsymbol{v}_h$ or $\boldsymbol{\mathsf{v}}_{r,2}$ to rotate $\boldsymbol{v}_t$.
In particular, the results degrade on both datasets when only utilizing either $\boldsymbol{\mathsf{v}}_{r,1}$ or $\boldsymbol{\mathsf{v}}_{r,2}$.
However, these two variants of QuatRE still outperforms QuatE, hence clearly showing the advantage of further using the relation-aware rotations in our QuatRE to enhance the correlations in knowledge graphs.

\section{Conclusion}
\label{sec:conclusion}

In this paper, we propose QuatRE -- a simple yet effective knowledge graph embedding model -- to learn the embeddings of entities and relations within the Quaternion space with the Hamilton product. 
QuatRE further utilizes two relation-aware rotations to strengthen the correlations between the head and tail entities.
Experimental results demonstrate that QuatRE outperforms up-to-date embedding models and produces state-of-the-art performances on well-known benchmark datasets for the knowledge graph completion task.


\bibliographystyle{ACM-Reference-Format}
\bibliography{references}

\end{document}